# Accelerating Grasp Learning via Pretraining with Coarse Affordance Maps of Objects

Yanxu Hou, Jun Li, *Senior Member*, IEEE

*Abstract*—Self-supervised grasp learning, i.e., learning to grasp by trial and error, has made great progress. However, it is still time-consuming to train such a model and also a challenge to apply it in practice. This work presents an accelerating method of robotic grasp learning via pretraining with coarse affordance maps of objects to be grasped based on a quite small dataset. A model generated through pre-training is harnessed as an initialization policy to warmly start grasp learning so as to guide a robot to capture more effective rewards at the beginning of training. An object in its coarse affordance map is annotated with a single key point and thereby, the burden of labeling is greatly alleviated. Extensive experiments in simulation and on a real robot are conducted to evaluate the proposed method. The simulation results show that it can significantly accelerate grasp learning by nearly three times over a vanilla Deep Q-Network -based method. Its test on a real UR3 robot shows that it reaches a grasp success rate of 89.5% via only 500 times of grasp tries within about two hours, which is four times faster than its competitor. In addition, it enjoys an outstanding generalization ability to grasp prior-unseen novel objects. It outperforms some existing methods and has the potential to directly apply to a robot for real-world grasp learning tasks.

## I. Introduction

LEARNING how to manipulate objects like human beings is a formidable challenge to robots. Grasping is one of the fundamental abilities that an intelligent robot should have. Moreover, robots are expected to have a certain level of learning competence to grasp prior-unseen novel objects in unstructured environments, e.g., in households and warehouses.

Deep learning makes this paradigm of robotic grasp learning possible. Recently, data-driven grasp learning has made significant progress. For example, learning to grasp either known or unseen objects according to their images instead of object models in [1]. Lenz *et al.* [2] train a convolutional neural network on a human-labeled dataset to predict grasp boundary boxes. They also verify that a robot can learn to grasp unseen objects. Generally, a large grasp dataset is necessary for the success of data-driven robotic grasp. But the preparation of a dataset such as, Cornell dataset [2], is time-consuming and laborious. Mahler *et al*. [3] try to alleviate heavy labeling burden by generating labeled data from synthetic point clouds.

Learning to grasp objects by trial and error is expected to achieve a good performance [4]–[10]. Particularly, Deep Reinforcement Learning (DRL) is quite suitable for grasp learning in unstructured environments. It can realize self-supervised grasp learning through interacting with environments, thus requiring no manual annotation of samples. However, they may suffer from low learning efficiency due to sparse or deceptive rewards. Moreover, random policies commonly performed in the early stages of training are even difficult to capture effective rewards [4]. Some work must train a few real robots and collect experiences concurrently for several weeks or months to achieve proper grasp policies. Obviously, it is too expensive to apply the methods in practice. Therefore, it is critically important to speed up a grasp learning process.

At present, accelerating grasp learning is a valued approach to obtain rapidly more effective grasp rewards, especially in early training stages. For example, Pinto and Gupta [4] employ an algorithm of background subtraction for identifying in advance regions of interest, which contains objects, to guide a robot's grasp in a correct position. Similarly, a method is proposed to assist grasping by using image recognition when grasping fails several times [10]. Although such techniques can capture more effective rewards in a simple background with a few apparent objects, they do not apply to the grasp of various objects in a sophisticated environment. Besides, they may easily get stuck at a local optimum or lose part of generalization ability. Some studies take advantage of heuristic initialization to obtain more learning rewards [6],[8]. A scripted policy is adopted to learn grasps by trial and error until it reaches a specific grasp success rate in [3]. Generally, learning from Demonstration (LfD) [11], i.e., imitation learning, can reduce the sample complexity of learning by trial and error. For example, Vecerik *et al.* [12] use both demonstrations and actual interactions to fill a replay buffer to speed up learning. But the demonstrations do not well deal with the difference in kinematics between users and robots and require the efforts of operation as well. It is intricate to acquire enough demonstrations, such as VR [13], teleoperation, and kinesthetic teaching. Insufficient generalization ability is another inherent drawback of LfD. In a word, the latter two approaches can speed up moderately grasp learning from different aspects at the cost of generalization ability or requiring manual efforts.

This work was supported in part by the National Natural Science Foundation of China under Grant 61773115, and Shenzhen Fundamental Research Program under Grant JCYJ20190813152401690. (Corresponding author: Jun Li.)
Y. Hou and J. Li are with the Ministry of Education Key Laboratory of Measurement and Control of CSE, Southeast University, Nanjing 210096, China (e-mail: yxhou@seu.edu.cn; j.li@seu.edu.cn).

On the other hand, simulation training instead of real one can reduce the real training stress. However, grasping performances suffer from the domain shift problem due to the gap between real and virtual scenarios [14].

To date, no efficient pre-training method exists for speeding up robotic grasp learning. In the field of deep learning-based object detection, a practical approach to learning acceleration leverages a preliminary model pre-trained on a large-scale image classification dataset, e.g., ImageNet. Unfortunately, it cannot be applied straightforward into speeding up grasp learning that commonly harnesses affordance maps of objects to be grasped. A reason is that grasp learning and image classification tasks have different goals, thereby often resulting in incompatible pre-trained models. For example, Zeng and Song *et al*. [10] harness an initial model pre-trained on ImageNet for promoting the value function approximator of Deep Q-Network (DQN). Nevertheless, the results are inferior even than learning from scratch in term of learning efficiency. In affordance map-based grasp learning, the affordance of an object is segmented for robotic manipulation. Affordance map-based grasp learning is like affordance segmentation in [15]. However, it must produce more fine-grained pixel-wise affordance referring to the probability of grasp success. Sawatzky *et al*. [16] train a fully convolutional network (FCN) from an object's image with few annotations of key points to generate the corresponding affordance map in a weakly supervised learning fashion. This approach not only achieves high accuracy of affordance segmentation but also significantly alleviates the burden of manual annotation. To the authors' best knowledge, the sample efficiency of weakly supervised learning is higher than RL's. Motivated by it, we present an accelerated grasp learning method via pretraining with coarse affordance maps of objects to be grasped.

The coarse affordance maps are generated in a way like that in [15] but using a different loss function. It makes the pre-trained model with coarse affordance maps, working as an initialization policy of DRL, well compatible with affordance-based grasp learning. Particularly, objects in input images are roughly annotated with few key points, and then the corresponding two-dimensional Gaussian distributions, i.e., affordances, are generated around the key points as the learning labels. Note that this technique is applicable to not only the DRL approaches like DQN but also any self-supervised affordance-based grasp learning by properly extension. The samples in our methods are composed of two parts, i.e., the roughly annotated images, and those sampled by trial and error. The former with less priori knowledge as inputs of pretraining contribute to training efficiency significantly. The latter leads to strong model generalizations due to samples' diversity.

The contributions of this work are:

1) An effective method is proposed to speed up robotic grasp learning by pretraining an FCN model with coarse affordance maps of objects on a quite small dataset. Our method can significantly speed up robotic grasp learning, especially in the early training stage. It has potential to directly apply into grasp learning in the real world, requiring no long training in simulation or real world. Only a key point is annotated on the affordance map of an object. It enjoys the higher sample efficiency and alleviates greatly the burden of sample labeling, compared to imitation learning and grasp detection [2]; and

2) Extensive experiments are conducted to evaluate our accelerated grasp learning method in both simulation and real environment. Specifically, a UR3 robot can grasp daily items with a mean success rate of 89.5% after 500 grasp tries within about 2 hours. Also, our method is of strong model generalization for grasping novel objects.

The problem of self-supervised grasp learning is formulated in Section II. Our method is presented in detail in Section III. Experimental results are given in Section IV. Section V concludes the work.

## II. PROBLEM FORMULATION

We consider a top-down parallel-jaw grasp on a plane. Self-supervised grasp learning is usually formulated as a Markov Decision Process $\langle S, A, r, \rho, \rho_0, T \rangle$, where $S$ is a state space, $A$ is an action space, $r: S \times A \to \mathbb{R}$ is an immediate reward, $\rho$ is a transition function, $\rho_0$ is the probability of initial state $s_0$, and $T$ is a fixed number of steps. An agent executes action $a_t$ in state $s_t$ at time $t$ following a policy $\pi(a_t|s_t)$, resulting in a new state $s_{t+1}$ and an immediate reward $r(s_t, a_t, s_{t+1})$. An episode terminates after $T$ steps or once a defined terminal state is reached.

We define a top-down parallel-jaw grasp as $a_t = (x, y, z, \alpha)$, where $x, y$, and $z$ denote the Cartesian coordinates of the grasp point, and $\alpha \in [0, \pi)$ represents the grasp orientations of a gripper in the yaw direction. In practice, we simplify grasp configuration by making $\alpha$ be discretized into 16 fixed orientations and $z$ be determined by the depth at $(x, y)$. The orientations can be equivalent to the rotation of input state $s_t$, i.e., a heightmap image. The action space $A$ is given by,

$$A = \left\{ (x, y, \alpha) \Big| x \in [0, H], y \in [0, H], \alpha \in \left\{ \frac{\pi}{8} \times i \Big| i = 0,1,2, \ldots, 15 \right\} \right\} \quad (1)$$

In Eq. (1), $H$ $W$ represents the height and width of input images, respectively. The original color and depth images of objects to be grasp, $I_{RGB}$ and $I_D$, captured by a fixed camera at time $t$, form a color heightmap $I'_{RGB}$ and a depth heightmap $I'_D$ by orthographically back-projecting upwards in the gravity direction. State space $S = \{s_t = (I'_{RGB}, I'_D) \mid I'_{RGB} \in \mathbb{R}^2, I'_D \in \mathbb{R}^2\}$. The reward function is represented intuitively as $r_{a_t}(s_t, s_{t+1}) = 1$, if a grasp succeeds; otherwise, $r_{a_t}(s_t, s_{t+1}) = 0$. The goal of grasp learning is to find an optimal grasp policy $\pi^*$ to maximize expected returns

$$\mathbb{E}_{a \sim \pi(a|s)} \left[ \sum_{t=0}^{T-1} \gamma^t r_{a_t}(s_t, s_{t+1}) \right],$$

where $\gamma \in [0,1]$ is a future discount rate.

## III. Proposed Method

We first pretrain an FCN model to generate coarse-grained affordance maps. Then the pre-trained model is used to initialize a Q-network. Finally, we train DQN-based grasp learning. The process is realized in Algorithm 1.

### A. Generation of Coarse Affordance Maps

Grasp affordance maps refer to the grasp success probability of all pixels in a captured image. In a self-supervised grasp learning paradigm, a robot can only obtain a one-pixel label of an image for a grasp attempt. Obviously, it leads to a low learning efficiency due to sparse and deceptive rewards that may exist. Generally, one can annotate each pixel of an image. For instance, the labels are obtained by fine annotating the graspable area in an image [14]. Definitely, it can increase the burden of labeling. In this work, we train an FCN model to generate coarse affordance maps. The pre-trained model is used to warmly start grasp learning to accelerate a self-supervised learning process. The dataset used in the pretraining stage is labeled by key points before self-supervised learning. It contributes to alleviating the burden of labeling.

*1) Dataset Preparation*

A sample $d_i$ of pretraining dataset $D = \{d_i\}, i = 1,2, \dots, K$, is obtained in the following steps, where $K$ is the number of samples in $D$. First, the objects are randomly placed into the workspace and its $I'_{RGB}$ and $I'_D$ are obtained by preprocessing the captured images. Second, the geometric center of an object, $(V, U) = \{(v_i, u_i)\}, i = 1,2, \dots, m$, is roughly annotated by key points in $I'_{RGB}$, where $m$ is the number of objects in $I'_{RGB}$, and $v_i$ and $u_i$ are the horizontal and vertical coordinates of the $i$th objects, accordingly. Note that, the accurate geometric center of an object is unnecessary. As shown in Fig. 1(a), we place four types of daily items into the workplace at random and annotate key points in yellow at the geometric center of objects roughly. Finally, a two-dimensional Gaussian distribution $g(v_i, u_i)$ called affordance maps is generated at the label point set $(V, U)$. Generally, the grasp success rate at different points of objects is

**Algorithm 1**: Grasp Learning Accelerating via Pretraining with Coarse Affordance Maps of Objects

1. Input: A FCN $Q_s(\theta_s)$ with a random initialization, hyperparameters of $Q_s(\theta_s)$ and Q-network $Q(s_t, a_t; \theta)$, and the number of collecting samples $K$.
2. Initialize an empty set $D = \emptyset$
3. For $i = 1,2, \dots, K$:
4.     Randomly place objects into the workspace.
5.     Capture original color and depth images, $I_{RGB}$ and $I_D$, back-project $I_{RGB}$ and $I_D$, a color heightmap $I'_{RGB}$ and a depth heightmap $I'_D$.
6.     Manually annotate every object's center $(V, U)$ in $I'_{RGB}$ with a key point.
7.     Generate a coarse affordance map of objects $Y_{GT}$ by Eq.(2) and Eq.(3). //see Section IV-A-1)
8.     $D = D \cup (I'_{RGB}, Y_{GT})$.//Append sample $(I'_{RGB}, Y_{GT})$ to $D$.
9. Train $Q_s(\theta_s)$ with SGD optimizer on $D$. //see Section IV-A-2)
10. Set parameters of $Q_s(\theta_s)$ as initialization parameters of $Q(s_t, a_t; \theta)$.
11. Obtain optimal grasp policy $\pi^*(a_t|s_t)$ by updating $Q(s_t, a_t; \theta)$. //see Section IV-C
12. Output: $\pi^*(a_t|s_t)$

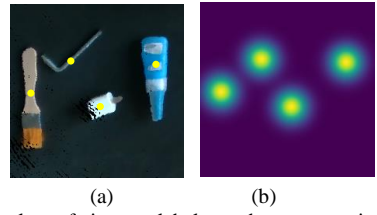

Fig. 1. Examples of image labels and coarse-grained affordances generated. (a) Manual labels of key points in yellow, and its (b) coarse grained affordance map with bright spots representing ground truths.

not the same. Therefore, we annotate object images by using two-dimensional Gaussian distribution instead of pixel-level category labels used in [14].

$$y^i_{GT} = g(v_i, u_i) = \frac{1}{2\pi\sigma_i^2} e^{-\frac{(v_i^2 + u_i^2)}{2\sigma_i^2}} \quad (2)$$

$$Y_{GT} = norm(y^1_{GT} + y^2_{GT} + \cdots + y^m_{GT}) \quad (3)$$

In Eq. (2), $\sigma_i$ is the distance to the boundary of object $i$ in $g(v_i, u_i)$, and $norm(\cdot)$ is a normalized operator in Eq. (3). The setting of $\sigma_i$ only needs to make in principle $y^i_{GT}$ cover object $i$ roughly at point $(v_i, u_i)$. Examples of $y^i_{GT}$ are illustrated in Fig.1(b) as bright spots. In practice, $\sigma_i$ can be set to the distance from the edge of object $i$ to the annotated point. $Y_{GT}$ is formed by normalized $y^i_{GT}$ and can be considered as a coarse affordance map of objects, as shown in Fig. 1(b). $Y_{GT}$ is the ground truth in the pretraining phase, and is of the same shape as $I'_{RGB}$. $Y_{GT}(x, y)$ represents the possibility of grasping at point $(x, y)$. The closer $Y_{GT}(x, y)$ is to 1, the more likely there is an object to be grasped at point $(x, y)$. Generating samples in this way can achieve strong robustness. Specifically, it allows that the annotated points deviate slightly from the center of an object, if the generated Gaussian distribution covers the target objects approximately.

*2) Generating Coarse-Grained Grasp Affordance Maps*

An FCN $Q_s(\theta_s)$ with parameters $\theta_s$ is adopted and trained on dataset $D$. It aims at predicting a coarse-grained grasp affordance map $Y_{Pred}$ for an input image. First, two DenseNet-121 are used in turn to extract the features of the color and depth heightmaps, respectively. Then, these feature maps are concatenated channel-wisely. Sequentially, two additional 1×1 convolutional layers are added with interleaved nonlinear activation functions (ReLU). Finally, $Q_s(\theta_s)$ outputs a coarse-grained affordance map of the same size as the input image by bilinearly upsampling.

*3) Loss Function*

We employ Kullback-Leiber Divergence (KL-Div) loss instead of Cross-Entropy loss in [14] to train $Q_s(\theta_s)$, given by

$$L(Y_{GT}, Y_{Pred}) = D_{KL}(Y_{GT} || Y_{Pred}) = \sum Y_{GT} \log \frac{Y_{GT}}{Y_{Pred}}, \quad (4)$$

where $L(Y_{GT}, Y_{Pred})$ is the KL-Div loss function. Cross-entropy loss is usually used in image segmentation that outputs the category of each pixel in an input image. Therefore, it is only used to distinguish graspable areas from ungraspable ones in affordance learning. Consequently, it leads to the same grasp success rate in all the graspable areas. KL-Div is often used to measure the similarity between two distributions. In view of grasp affordance maps as an expression of the grasping probability distribution, we adopt KL-Div loss to allow $Q_s(\theta_s)$ directly inferring the grasp success rate for each pixel in the image, namely, approximating to $Y_{GT}$ by predicating $Y_{Pred}$. A comparative experiment on training $Q_s(\theta_s)$ with different loss functions is conducted in Section IV.

### B. Warm Start of Grasp Learning

We leverage the pre-trained $Q_s(\theta_s)$ to speed up grasp learning instead of learning from scratch. In particular, a small-scale dataset is prepared first in real world, and then $Q_s(\theta_s)$ is trained on the dataset. Next, the parameters of pre-trained $Q_s(\theta_s)$ are considered as the initialization parameters of $Q(s_t, a_t; \theta)$ for warmly starting grasp learning. $Q(s_t, a_t; \theta)$ is the Q-network in DQN. It is more compatible with pre-trained $Q_s(\theta_s)$ than that directly pertained on a large-scale image dataset for robotic grasp learning. The goal of training $Q_s(\theta_s)$ is identical to grasp learning by trial and error. Both goals are to segment affordances from image inputs. Consequently, utilizing the pre-trained $Q_s(\theta_s)$ to warmly start grasp learning can guide robotic grasping at a proper position in early training stages. It can dramatically accelerate the learning process.

### C. DQN-based Grasp Learning

Optimal grasp policy $\pi^*(a_t|s_t)$ is trained by a classical value-based DRL algorithm, i.e., DQN. $Q(s_t, a_t; \theta)$ is an approximator of the action-value function of DQN in which a duplicate operation and a rotation transformation are added in turn in front of the backbone of $Q_s(\theta_s)$. $Q(s_t, a_t; \theta)$ outputs grasp affordance maps $C = [c_0, c_1, \ldots c_i, \ldots, c_{15}] \in \mathbb{R}^{16 \times H \times W}$, where $c_i$ represents the grasp affordance map in the $i$ th orientation. An optimal grasp action $a_t^*$ is selected according to $C$, i.e., $a_t^* = (x^*, y^*, \alpha^*) = \underset{x^*, y^*, \alpha^*}{\mathrm{argmax}}\, C(x, y, \alpha)$. $Q(s_t, a_t; \theta)$ is updated by a smooth L1 loss function. After updating $Q(s_t, a_t; \theta)$ for a reasonable number of steps, an optimal grasp policy $\pi^*(a_t|s_t) = \underset{\theta}{\mathrm{argmax}}\, Q(s_t, a_t; \theta)$ is obtained for a robot.

In summary, unlike Vanilla DQN-based method learning from scratch, our method can accelerate grasp learning by leveraging a pretraining technology.

### D. Training Details

$Q_s(\theta_s)$ is trained by using a Stochastic Gradient Descent optimizer with a momentum at learning rate 10e-4, momentum 0.9, and weight decay 2e-5. It is trained on a computing server with an NVIDIA 1080Ti GPU and an Intel Xeon CPU E5-2620 v3 @ 2.4Ghz on PyTorch.

First, the parameters of pre-trained $Q_s(\theta_s)$ are loaded into $Q(s_t, a_t; \theta)$. Then, perform the following steps repeatedly. $n$ objects are selected at random and placed into the workspace. The hyper parameters of $Q(s_t, a_t; \theta)$ are configured as same as those of $Q_s(\theta_s)$. The replaying technology of priori experience is injected in DQN. The exploration strategy is ε-greedy with $\epsilon$=0.5 initially and then decayed to 0.001 in run time. The future discount factor $\gamma$=0.9.

## IV. PERFORMANCE EVALUATION

We test our method in both simulation environment and real world. First, we qualitatively analyze the effects of using KL-Div loss on generation of grasp affordance maps. Second, our method is compared with the existing acceleration methods in terms of acceleration performances. Finally, we evaluate our method on a UR3 robot in the real world.

### A. Setup of Simulation Environment

A virtual UR5 robotic arm mounted with an RG2 gripper is constructed in the robot simulation platform V-rep, as shown in Fig. 2. The motion planning of the arm is linearly interpolated. We choose a total of 9 different types of 3D objects and their colors, quantity, and shapes are generated at random.

### B. Comparison of Loss Functions

Mean Square Errors (MSE), Smooth L1, and KL-Div loss are employed as loss functions to train the model for generating coarse affordance maps on a small dataset, respectively. The dataset contains 50 images and is constructed according to the method of dataset preparation presented in Section III. Except for loss functions, the network structure, hyperparameter, and optimizers are the same as those in Section III-A. MSE is a commonly-applied criterion in regression problems. Generally, Smooth L1 can remove the influence of outliers compared with MSE. The results of comparison are shown in Fig. 3, where all values are normalized to [0, 1].

The value of a pixel in Fig.3 indicates the grasp success rate at the pixel. The greater the value is, the more successful the grasp is. According to Fig. 3(a) and (d), we find that using KL-Div loss has the following advantages. First, the pixel values in the non-objects area tend to be zero, representing seldom grasps occurring in a non-object region. Second, the generated affordance can cover the object completely, and a robotic grasp has a greater success rate near the center of an object. MSE and smooth L1 loss suffer from more noise in the non-object region. It indicates wrong grasps in non-object regions. Therefore, we select KL-Div as the loss function in generating of coarse affordance maps.

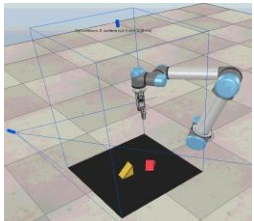

Fig. 2. Simulation scenario in V-rep

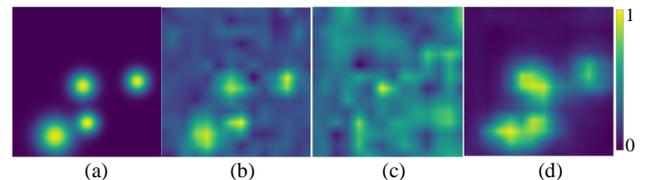

(a)      (b)      (c)      (d)

Fig. 3. Predicted results of $Q_s(\theta_s)$ by using different loss functions, (a) Label, (b) Mean Square Error, (c) Smooth L1, (d) KL-Div.

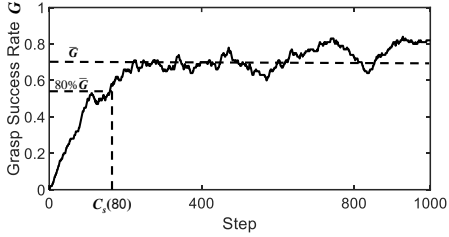

Fig. 4. Schematic diagram of $\bar{G}$ and $C_s(p)$.

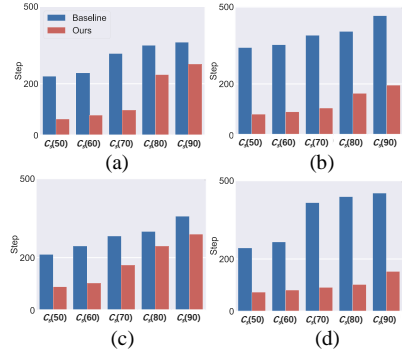

Fig.6. Histograms of two methods at different $C_s$ with (a) 2 objects, (b) 4 objects, (c) 6 objects, and (d) 10 objects.

### C. Quantitative Analysis of Acceleration Performance

#### 1) Evaluation Metrics

Grasp Success Rate, $G = S/M$ defined in [7] is used to measure the performance of self-supervised grasp learning, where $S$ is the number of successful grasps within $M$ times of grasps. Unless otherwise stated, $M = 100$. As shown in Fig. 4, the convergence curves of $G$ fluctuate severely. It is not appropriate to use a single value to refer to a final stable Grasp Success Rate. Therefore, we give a new definition of $G$.

**Definition 1**: Consider a sequence of $G$s, denoted by $y(i), i = 1,2, \dots, n$. If $|y(i) - y(i-1)| < \Delta$, is satisfied successively in $k$ times, where $\Delta$ and $k$ are the thresholds configured by human priori knowledge. A *stable Grasp Success Rate* ($\bar{G}$) is given by,

$$\bar{G} = \frac{1}{k}\sum_{i=n-k}^{n} y(i), \quad (6)$$

where $y(i) = H_i/L$ represents $H_i$ times of successful grasps in $L$ times of grasps of the $i$th sliding box and $n = T/L$ with $T$ representing total training iteration count.

$\bar{G}$ can effectively resolve the fluctuation issue arising from the use of a single $G$. Accordingly, we define an indicator of convergence steps ($C_s$).

**Definition 2**: $C_s$ is the number of steps of $G$ needed to converge to $\bar{G}$. Generally, $C_s(p)$ is the number of steps of $G$ needed to converge to $p\%$ of $\bar{G}$.

#### 2) Results and Analysis

We compare first quantitatively the acceleration performance of our method with the baseline, i.e., the vanilla DQN-based grasp learning shown in Section III-C. In our simulation, 2, 4, 6, and 10 objects are randomly placed in turn in the workspace to reflect the different object sparsity. We use separately our method and the baseline to learn a grasp policy by training 1000 times.

The resultant $G$ curves with 95% confidence interval computed according to 8 groups of experiments are shown in Fig. 5. The red curve of grasp success rates obtained by our method is obviously more efficient than the blue curve of grasp success rates generated by the baseline method. Specifically, our method reaches nearly 60% $G$ in about 200 training steps, which is nearly 3 times faster than the baseline. When a different number of objects are initially placed in a workspace, the variance of $G$ of our method is significantly smaller than that of the baseline. It shows that our $G$ has little relevance to the object sparsity. Therefore, it shows that our method can effectively accelerate grasp learning despite multiple objects to be grasped in a workspace.

Figure 6 shows $C_s(50)$, $C_s(60)$, …, $C_s(90)$ of the baseline and our method given different numbers of objects to be grasped. Clearly, our method requires fewer tries than the baseline does to achieve the same $\bar{G}$ in all cases. Moreover, we calculate the **acceleration ratio** of our method to the baseline via,

$$m = \frac{r - r'}{r'}, \quad (7)$$

where $r$ and $r'$ are the $C_s(p)$ of the baseline and our method, respectively. The resultant acceleration ratios are listed in Table I.

TABLE I.    ACCELERATION RATIOS OF OUR METHOD TO THE BASELINE

|  | $C_s(50)$ | $C_s(60)$ | $C_s(70)$ | $C_s(80)$ | $C_s(90)$ |
|---|---|---|---|---|---|
| 2 obj | 2.63 | 2.10 | 2.16 | 0.48 | 0.31 |
| 4 obj | 3.15 | 2.86 | 2.66 | 1.47 | 1.38 |
| 6 obj | 1.38 | 1.39 | 0.65 | 0.23 | 0.23 |
| 10 obj | 2.31 | 2.27 | 3.56 | 3.29 | 1.97 |

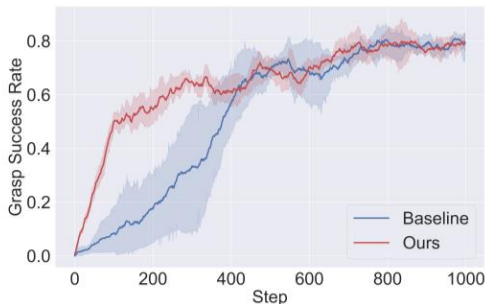

Fig. 5. The performance of our methods compared to baseline.

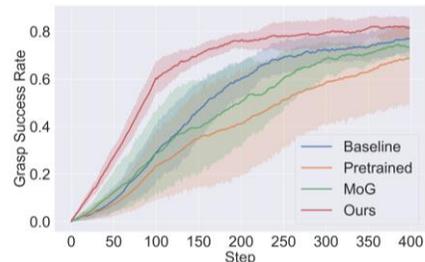

Fig.7. Comparison of accelerated methods in simulation.

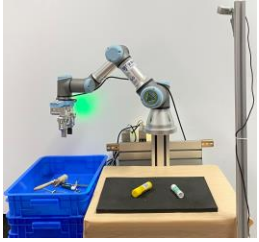

Fig.8. UR3 Experimental Scenario in real world.

Specifically, first, all values in Table I are greater than 0. It suggests that our method can efficiently speed up the baseline method in all cases. Second, $m$ is greater than 1 and even reach to 3.15 in the early stage such as $C_s(50)$ and $C_s(60)$ with four objects to be grasped. It means that our method is 1~3 times faster than the baseline. Third, $m$ decreases gradually and then tends to be stable during the training.

Moreover, we compare our method with the following methods: an approach of using Mixture of Gaussians(**MoG**) in [4] and a method with **P**re-training backbone network **o**n **I**mageNet(**PoI**) in [10]. Each experiment runs 10 rounds with different initial seeds and only updates 400 times in the early stage of training.

The results are shown in Fig. 7. We find that our method can converge significantly faster and shows greater utilization of samples than all other methods as shown in Fig. 7. It is attributed to our method's warm start of grasp learning. Specifically, our method guides a robotic grasp at a proper position and effectively obtains the rewards in the initial stage. It can achieve nearly 80% $G$ by training merely about 200 steps. Remarkable, our method is robust to different scenarios and can quickly converge in all instances. As shown in Fig. 8, MoG and PoI methods have more volatility than ours and only accelerate the learning process in a few cases.

### D. Grasp Learning of a Real Robot

Two experiments are conducted in order to evaluate whether our method can be directly applied into a real robot in terms of accelerated performance and generalization of grasping novel objects. Both performances play a crucial role in practice. The experimental system consists of a UR3 robot equipped with a parallel-jaw and a RealSense 435 RGB-D camera, as shown in Fig. 8. The video of the UR3 robot's grasp learning experiments are displayed in the Supplementary File.

Unlike the primitive geometries grasped in a simulation environment, we select 12 kinds daily items of irregular shapes with complicated shapes, colors, and material, as shown in Fig. 9(a). It needs more accuracy in robotic grasping positions and orientations.

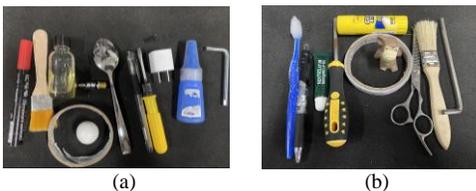

(a)          (b)

Fig. 9. Daily items for real robotic experienment. (a) 12 daily items of learning to grasp and (b) 10 unseen novel objects.

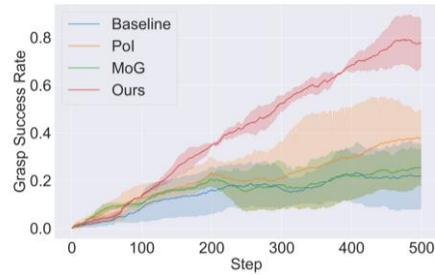

Fig.10. Comparing with other methods in real world.

Twenty images containing several arbitrarily placed items are captured to construct a small dataset. Each image shows several arbitrarily placed items, as illustrated in Fig. 1(a). Then, the method in Section III-A is adopted to train a model to generate a coarse affordance map. Then, the pre-trained model is harnessed as an initialization policy to warmly start grasp learning. The real grasp learning process of the real robot is like that in simulation. UR3 robot attempts to grasp items randomly placed in the workspace and updates $Q(s_t, a_t; \theta)$. Repeat the above process, until all the items in the workspace are grasped successfully. We conduct the experimental process for 3 rounds by using MoG, PoI, the baseline, and our method. The $G$ curves are plotted with a 95% confidence interval, as shown in Fig. 10.

It shows that the $G$ curve of our method in red is dramatically faster than that of its peers. Our method reaches 79.14% average $G$ after 3 runs of experiments in 500 steps. Our method is nearly 3 times faster than the baseline which barely reaches about 21% $G$ in 500 steps. By observation of the training process, we find that our method more likely to grasp successfully the items at the very beginning of training. It indicates that our method can capture effective rewards more effortlessly. Remarkably, our method can achieve up to 89.56% $G$ in less than 500 steps, as illustrated in Fig.10. The baseline method needs 2500 steps to reach about 80% $G$ in [10]. In a word, our method can significantly improve the efficiency of grasp learning and outperforms other acceleration methods through the same number of training steps.

In addition, we compare our method with other three acceleration methods by grasping ten novel items. as shown in Fig. 9(b), including items like the training data, e.g., a new scissor, brush, and screw driver, and completely distinctive ones, i.e., a monkey doll. 10 novel items are randomly placed in the workspace. The experiment is repeated for 8 rounds and each round attempts to grasp 20 times.

TABLE II.     COMPARING $G$ OF GRASPING NOVEL OBJECTS ON DIFFERENT METHOD

|   | Baseline | Ours | MoG | PoI |
|---|---|---|---|---|
| 1 | 0.12 | 0.76 | 0.32 | 0.33 |
| 2 | 0.67 | 0.70 | 0.28 | 0.28 |
| 3 | 0.22 | 0.66 | 0.33 | 0.34 |
| 4 | 0.18 | 0.65 | 0.21 | 0.22 |
| 5 | 0.13 | 0.61 | 0.18 | 0.19 |
| 6 | 0.71 | 0.72 | 0.29 | 0.41 |
| 7 | 0.59 | 0.81 | 0.27 | 0.21 |
| 8 | 0.49 | 0.76 | 0.41 | 0.33 |
| Mean | 0.3887 | 0.7087 | 0.2865 | 0.2787 |
| ±Std | ±0.2356 | ±0.0625 | ± 0.0668 | ±0.0711 |

Table II shows that $G$ of grasping novel items of the models trained in order by our method, the baseline, MoG, and PoI for 500 steps. The experimental results reveal that our method achieves 70.87% $G$ and significantly outperforms its three peers in the same number of steps. It shows that our method has an outstanding generalization ability.

In summary, the real experiments on a UR3 robot reveal that our method can achieve nearly 89.56% $G$ via less than 500 grasping attempts, and possesses a fine generalization ability of grasping novel objects. It is promising to directly apply our method into real robotic grasp learning.

## V. Conclusion

In this work, we present an accelerating method of robotic grasp learning via pre-training with coarse affordance maps of objects. It only requires to use a small set of samples for pretraining a model to generate coarse affordances maps. All efforts one needs to pay are merely to label every sample with one key point, roughly at the center of an object. It can greatly alleviate the burden of labelling. Simulation and real experiments show that our method can accelerate grasp learning significantly, compared with the vanilla DQN-based, MoG, and PoI. Using our method, a high grasp success rate within only hundreds of grasping attempts can be achieved. Our method can avoid transferring from the virtual to real, and does not require learning from scratch with massive samples. It provides a potential solution to carry grasp learning in real world. Our ongoing work is to extend our method of grasp learning in a discrete action space to some challenging task learning in a continuous action space.